\documentclass{iaf}
\usepackage[utf8]{inputenc}
\usepackage{times,latexsym,graphicx,xspace,stmaryrd,subfig,yhmath,color,eepic}
\usepackage{amsfonts}
\usepackage{amsmath}
\usepackage{rotating}
\usepackage{float}

\usepackage{MnSymbol}
\DeclareSymbolFont{Symbols}{OMS}{cmsy}{m}{n}
\DeclareMathSymbol{\emptyset}{\mathord}{Symbols}{"3B}

\def\indu{\ifmmode \mathcal{INDU} \else $\mathcal{INDU}$\xspace \fi}

\def\wftimeofeventfunction{\Theta}

\def\wfrepetitionoperator{\uparrow}
\def\wfsequenceoperator{\to}
\def\wfinterpretationdomain{\Omega}

\def\wfinterpretationfunction{\wfinterpretationof{\cdot}}
\def\extwfinterpretationfunction{\extwfinterpretationof{\cdot}}
\newcommand\wfinterpretationof[1]{\left\llbracket#1\right\rrbracket}
\newcommand\extwfinterpretationof[1]{\left\llbracket#1\right\rrbracket}
\newcommand\wfloop[1]{{\rcurvearrowdown}#1{\rcurvearrowup}}
\newcommand\miniwfconj[2]{\genfrac{[}{]}{0mm}{}{\displaystyle#1}{\displaystyle#2}  }

\newcommand\wfconj[2]{\left[\begin{array}{c}#1\vspace{1mm}\\#2\end{array}\right]}
\newcommand\miniwfdisj[2]{  \genfrac{\langle}{\rangle}{0mm}{}{\displaystyle #1}{ \displaystyle#2}  }

\newcommand\wfdisj[2]{\left\langle\begin{array}{c}#1\vspace{1mm}\\#2\end{array}\right\rangle}
\newcommand\wfsequence[2]{#1\wfsequenceoperator#2}
\newcommand\wfpsequence[2]{\left(#1\wfsequenceoperator#2\right)}
\newcommand\wftimeofevent[1]{\wftimeofeventfunction(#1)}
\newcommand\wfrepetition[2]{#1\wfrepetitionoperator#2}

\newcommand\miniwfcdisj[3]{\left\langle #1 \genfrac{}{}{0mm}{}{\displaystyle#2}{\displaystyle#3}\right\rangle }

\def\induinterpretationfunction{\induinterpretationof{\cdot}}
\newcommand\induinterpretationof[1]{\left\llbracket#1\right\rrbracket}
\def\intervaldomain{\mathbb{Q}}

\def\proptrue{\mathtt{vrai}}
\def\propfalse{\mathtt{faux}}
\def\proptruthvalues{\{\propfalse,\proptrue\}}

\newcommand\conddef[4]{
	\left\{
		\begin{array}{c l}
			#1 & \quad #2 \\
     		#3 & \quad #4
  		\end{array}
  	\right.
}

\def\tand{\text{ et }}
\def\tor{\text{ ou }}
\def\tif{\text{ si }}
\def\totherwise{\text{ sinon }}

\newcommand\alrel[1]{\mathrel{\{\textsf{#1}\}}}
\newcommand\alrelnb[1]{\ifmmode \mathrel{\textsf{#1}} \else \textsf{#1}\xspace \fi}

\def\wfqcninject{\mathcal{R}}

\def\wfint{\interpretation}
\def\qaint{\interpretation}
\def\ewfint{\interpretation}
\def\interpretation{\mathfrak{m}}
\def\qainv{\ifmmode ^\smile \else $^\smile$\xspace \fi}

\usepackage{tikz}
\usetikzlibrary{arrows}
\def\subsumepar{\sqsubseteq}
\def\equivalenta{\mathrel{{\subsumepar}\!\!\!\text{\reflectbox{$\subsumepar$}}}} 
\def\langagewf{{\mathcal{W}}}
\def\langagewfe{{\mathcal{W}}_\mathfrak{B}}
\def\mlm#1{\text{\begin{tabular}{c}#1\end{tabular}}}

\def\nswf{\!\!\!\!\!} 

\newcommand\wfconjIIsepar[2]{\left[\nswf\mlm{$#1$\\[4mm] $#2$}\nswf\right]}
\newcommand\wfconjIII[3]{\left[\nswf\mlm{$#1$\\ $#2$\\ $#3$}\nswf\right]}
\newcommand\wfconjIV[4]{\left[\nswf\mlm{$#1$\\ $#2$\\ $#3$\\ $#4$}\nswf\right]}

\newcommand\wfdisjIIsepar[2]{\left\langle\nswf\mlm{$#1$\\[3mm] $#2$}\nswf\right\rangle}
\newcommand\wfdisjIII[3]{\left\langle\nswf\mlm{$#1$\\ $#2$\\ $#3$}\nswf\right\rangle}

\newcommand\wfsequenceIII[3]{#1\wfsequenceoperator#2\wfsequenceoperator#3}

\renewcommand\wfsequence[2]{#1{\wfsequenceoperator}#2}

\let\miniwfconj=\wfconjII
\let\miniwfdisj=\wfdisjII
\let\wfconj=\miniwfconj
\let\wfdisj=\miniwfdisj
\let\lien=\wfqcninject 

\def\QCN{{\mathcal{N}}}

\newcommand{\wf}{flux opérationnel\xspace}
\newcommand{\wfs}{flux opérationnels\xspace}
\newcommand{\Wf}{Flux opérationnel\xspace}
\newcommand{\Wfs}{Flux opérationnels\xspace}

\annee{2012}
\titre{Extension du formalisme des flux opérationnels par une algèbre temporelle}

\auteurs{Valmi Dufour-Lussier$^{1,2,3}$ \and Florence Le Ber$^{4,1}$ \and Jean Lieber$^{1,2,3}$} 
\institutions{
	$^1$ Université de Lorraine, LORIA UMR 7503, 54506 Vand\oe{}uvre-lès-Nancy \\
	$^2$ CNRS, 54506 Vand\oe{}uvre-lès-Nancy \qquad\qquad\qquad\qquad
	$^3$ Inria, 54602 Villers-lès-Nancy \\
	$^4$ LHYGES, Université de Strasbourg/ENGEES, CNRS, 67000 Strasbourg
}

\mels{\{valmi.dufour,florence.le-ber,jean.lieber\}@loria.fr}

\begin{document}
\creationEntete

\begin{resume} 
Les flux opérationnels \textit{(workflows)} constituent un important langage de représentation des connaissances sur les processus, mais sont également de plus en plus utilisés pour raisonner sur ce type de connaissances. En revanche, ils sont limités pour l'expression de contraintes temporelles entre activités. Les algèbres qualitatives d'intervalles peuvent représenter des relations temporelles plus fines, mais elles sont incapables de reproduire toutes les structures de contrôle des flux. Cet article définit une sémantique, fondée sur la théorie des modèles, commune aux flux opérationnels et aux algèbres d'intervalles, rendant possible l'inter-opérabilité de systèmes de raisonnement utilisant ces deux formalismes. Cela met également en évidence des propriétés et des possibilités d'inférences intéressantes, à la fois pour les flux opérationnels et pour les flux étendus par l'utilisation d'une algèbre qualitative. Finalement, nous discutons de formalismes similaires, proposant également une base théorique au formalisme des flux et étendant ce dernier.
\end{resume}

\begin{abstract}
	Workflows constitute an important language to represent knowledge about processes, but also increasingly to reason on such knowledge.
	On the other hand, there is a limit to which time constraints between activities can be expressed.
	Qualitative interval algebras can model processes using finer temporal relations, but they cannot reproduce all workflow patterns.
	This paper defines a common ground model-theoretical semantics for both workflows and interval algebras,
	making it possible for reasoning systems working with either to interoperate.
	Thanks to this, interesting properties and inferences can be defined, both on workflows and on an extended formalism combining workflows with interval algebras.
	Finally, similar formalisms proposing a sound formal basis for workflows and extending them are discussed.
\end{abstract}

\section{Introduction} 
La capacité des flux opérationnels (\emph{workflows})
 à décrire des processus temporels explique qu'ils sont
 de plus en plus utilisés, non seulement comme outils pour
 représenter des informations à propos de l'exécution de
 tâches complexes, mais aussi pour raisonner
 sur ces tâches
 (voir, par exemple,~\cite{hsu2011ie} et~\cite{minor10iccbr}).
Les flux opérationnels définissent un ordre sur les activités
 pour l'exécution d'une tâche et permettent l'utilisation de
 structures de contrôles plus complexes, mais ils ne permettent pas
 d'exprimer des relations temporelles plus fines entre ces
 activités.
À l'inverse, les algèbres qualitatives sur les intervalles
 peuvent être utilisées, avec des activités réifiées sous forme
 d'intervalles, afin d'exprimer des relations temporelles
 fines, mais elles ne permettent pas d'exprimer des structures
 de contrôle telles que les boucles ou la disjonction (i.e.
 le choix entre activités).

Le rapprochement entre ces deux approches n'est pas entièrement nouveau.
En particulier, leur combinaison a été ébauchée dans~\cite{lu06adc},
pour permettre la vérification de la cohérence des flux opérationnels.

Cet article propose un formalisme de flux opérationnels étendus
 qui intègre une algèbre qualitative au sein d'un formalisme
 de flux opérationnels afin d'étendre l'expressivité de chacun de ces
 formalismes.
Afin de rendre cela possible, un langage formel des flux opérationnels
 est défini et accompagné d'une sémantique en théorie des modèles
 d'un type qui peut aussi s'appliquer aux algèbres qualitatives.

La section~\ref{sec:wf} introduit une syntaxe définie inductivement
 pour les flux opérationnels.
Les algèbres qualitatives temporelles sont présentées dans la
 section~\ref{sec:qa}.
La section~\ref{sec:sem} propose des sémantiques en théorie des modèles
 qui sont
 similaires pour les flux opérationnels et les algèbres temporelles.
Cela permet de proposer un formalisme de flux opérationnels étendus
 par une algèbre des intervalles et une sémantique pour ce formalisme,
 ainsi que la section~\ref{sec:ext} le décrit.
La section~\ref{sec:disc} discute des travaux connexes
 et la section~\ref{sec:concl} conclut l'article et présente
 ses perspectives.

\section{Syntaxe des flux opérationnels}
\label{sec:wf}
Un \wf 
est la description formelle d'un processus. Les \wfs sont apparus avec la révolution industrielle, mais ils n'ont pris leur forme actuelle qu'à partir des années 1980, pour améliorer la productivité -- voir par exemple~\cite{en80acmcs,bp84tois}. L'objectif principal d'un \wf est d'organiser des activités d'un processus à l'aide d'un flux de contrôle. 

Les activités décrites peuvent l'être à différents niveaux de granularité : une activité contient d'autres activités. Ainsi un \wf complet peut être vu comme une seule activité de forte granularité.  Nous allons traiter par la suite toutes les activités comme des \wfs pouvant être combinés grâce à des opérations de contrôle pour créer des \wfs plus grands : ce point de vue compositionnel nous aide à définir un langage formel de description des \wfs. Parmi les activités, certaines sont considérées comme des briques de base non décomposables et nous les appelons \wfs atomiques.

\begin{figure}
  {
  \def\leve#1#2{\raisebox{#1mm}{#2}}
  \def\mef#1{{\huge$#1$}} 
  \def\aa{\mef{\varphi}}
  \def\bb{\mef{\chi}}
  \begin{center}
  \subfloat[Séquence.]{\label{fig:wf-seq}
      ~\leve{8}{\resizebox{0.08\textwidth}{!}{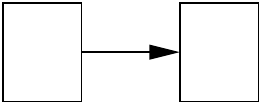}}~~}
  \subfloat[Conjonction.]{\label{fig:wf-conj}
      \,\resizebox{0.10\textwidth}{!}{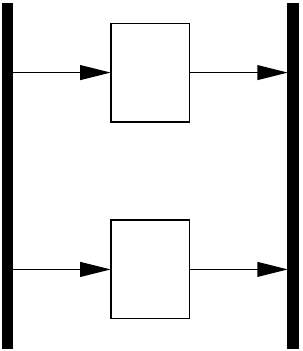}\,}
  \subfloat[Disjonction.]{\label{fig:wf-disj}
      \leve{3}{\resizebox{0.12\textwidth}{!}{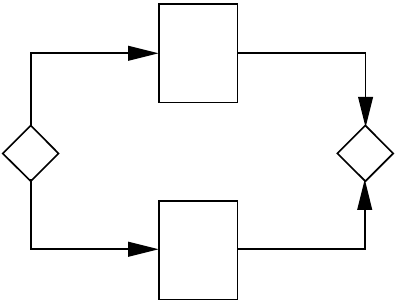}}}
  \subfloat[Boucle.]{\label{fig:wf-loop}
      \leve{5}{\resizebox{0.12\textwidth}{!}{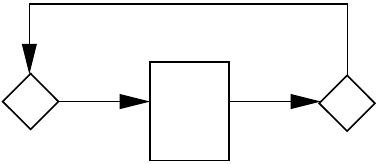}}}
  \end{center}
  }
\caption{Structures de contrôle de flux.}
	\label{fig:wf-cfs}
\end{figure}

L'opération de contrôle primaire dans un \wf est la séquence  (figure~\ref{fig:wf-seq}), qui signifie simplement qu'un \wf ne peut démarrer que quand un autre a été exécuté. Les autres opérations sont l'embranchement, qui permet une exécution concurrente de \wfs, la jointure, qui permet de les synchroniser, le choix,  qui force l'exécution exclusive d'un \wf parmi plusieurs, et la fusion qui conclut l'opération précédente~\cite{aalst03dpd}. L'embranchement et la jointure sont utilisés pour créer une structure de conjonction (figure~\ref{fig:wf-conj}),  tandis que les opérations de choix et de fusion sont utilisées pour créer une structure de  disjonction  (figure~\ref{fig:wf-disj}) ou une structure de boucle (figure~\ref{fig:wf-loop}).

\begin{figure}
   \begin{center}
\subfloat[Diagramme de flux UML.]{\label{fig:wf-uml}{%
    \def\mef#1{{\huge$#1$}} 
    \def\aa{\mef{\alpha}}
    \def\bb{\mef{\beta}}
    \def\cc{\mef{\gamma}}
    \def\dd{\mef{\delta}}
    \def\ee{\mef{\varepsilon}}
    \def\ff{\mef{\zeta}}
    \resizebox{0.4\textwidth}{!}{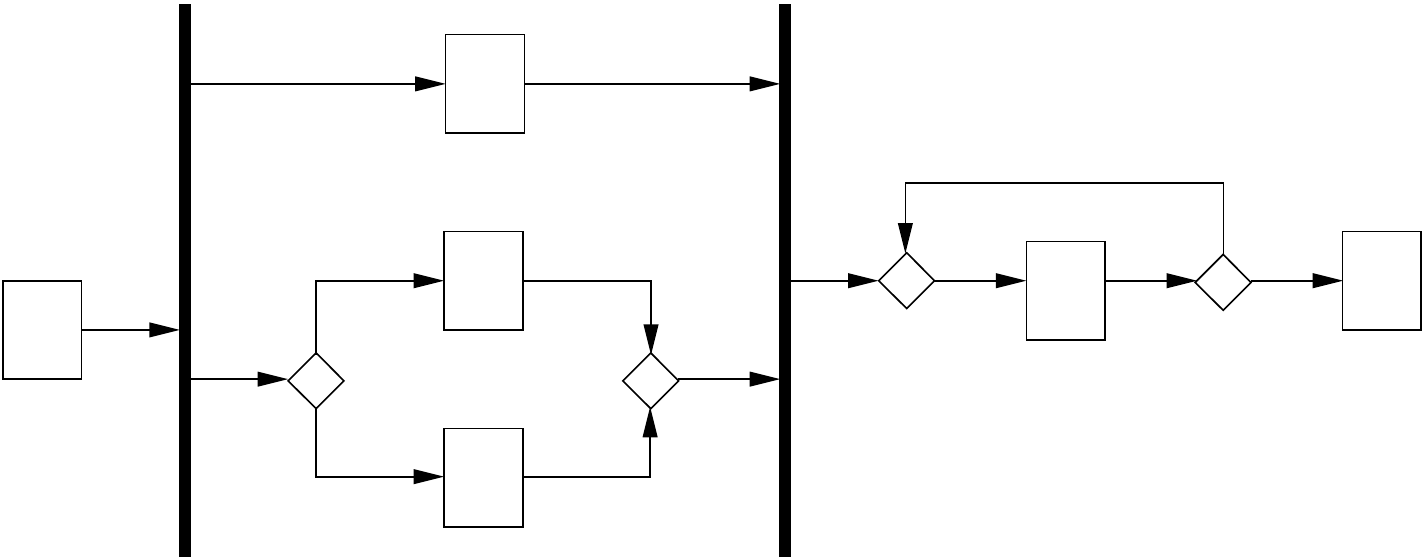}
    }
}
\\
  	\subfloat[Langage formel de  \wf.]{ \label{fig:wf-compact}
                        $ \qquad\qquad
                        	\displaystyle
                        		\wfsequence{
										\wfsequence{
										\wfsequence
										{\alpha}
				                  {\miniwfconj{ \beta }{ \miniwfdisj{ \gamma }{ \delta } }}}
                    				{\wfloop{ \varepsilon }}}
                    				{\zeta}
                        \qquad\qquad $
   	}
   \end{center}
   \caption{ Un exemple de \wf.}
   \label{fig:wf}
\end{figure}

La figure~\ref{fig:wf-uml} montre un exemple de \wf illustrant chacune des structures de contrôle, sous la forme d'un diagramme de flux UML, couramment utilisé pour représenter les \wfs.  Dans ce papier, nous allons définir un langage formel  $\mathcal{W}$ pour les \wfs qui utilise une représentation plus synthétique des activités et des opérations de contrôle de flux. Ce langage est construit de manière inductive, permettant ainsi une définition plus facile de sa syntaxe compositionnelle et de sa sémantique.

Un \wf modélise un ensemble d'activités qui doivent être réalisées pour accomplir une tâche donnée. Selon cette définition, toute partie d'un \wf est aussi un \wf. Un \wf que l'on ne peut pas décomposer en plus petits \wfs est un \wf atomique. Le langage des \wfs est le plus petit ensemble $\mathcal{W}$ construit à partir de l'ensemble  $\mathcal{A}$ des \wfs atomiques tel que :

\begin{description}
\item[\Wf atomique] Si $\alpha \in \mathcal{A}$, alors  $\alpha \in \mathcal{W}$;
\item[Séquence] Si $\varphi, \chi \in \mathcal{W}$, alors  $\wfsequence{\varphi}{\chi} \in \mathcal{W}$;
\item[Conjonction] Si $\varphi, \chi \in \mathcal{W}$, alors  $\miniwfconj{\varphi}{\chi}\in \mathcal{W}$;
\item[Disjonction] Si $\varphi, \chi \in \mathcal{W}$, alors  $\miniwfdisj{\varphi}{\chi} \in \mathcal{W}$;
\item[Boucle] Si $\varphi \in \mathcal{W}$, alors  $\wfloop{\varphi} \in \mathcal{W}$.
\end{description}
Dans la suite, les \wfs atomiques sont notés  $\alpha$, $\beta$, etc., éventuellement indexés, et les autres \wfs sont notés $\varphi$, $\chi$ et $\psi$, éventuellement indexés.
Une représentation du \wf de la figure~\ref{fig:wf-uml} dans le langage $\mathcal{W}$ est montrée à la figure~\ref{fig:wf-compact}.
 
Parmi les applications possibles des \wfs et des algèbres qualitatives -- où ils sont utilisés indépendamment pour représenter des connaissances incomplètes -- se trouve le domaine culinaire, qui a déjà été étudié tant du point de vue des flux~\cite{minor10ccc} que de celui des algèbres temporelles~\cite{lln10rte}. C'est ce domaine qui servira d'exemple. Les flux opérationnels sont utilisés pour modéliser de nombreux autres types de processus dont les activités peuvent présenter des interactions temporelles, par exemple les processus scientifiques~\cite{ludaescher06ccpe} et les processus d'affaires~\cite{aalst03icbpm}.

La figure~\ref{fig:extxt} décrit le texte d'une recette qui est représentée, autant que possible, par le \wf de la figure~\ref{fig:exwf}: on voit que le langage
 $\mathcal{W}$ n'est pas assez expressif pour exprimer la connaissance que <<\,saisir le foie gras doit se faire $30$ secondes avant que le tournedos
 ne soit cuit de sorte que les deux ingrédients soient prêts au même moment\,>>, ni le terme <<\,immédiatement\,>>.
 
 \begin{figure*}
	\begin{center}
		\subfloat[Texte d'une recette.]{ \label{fig:extxt}
					\fbox{
					\begin{minipage}{.75\linewidth}
					Faire frire le tournedos dans le beurre. 
30 secondes avant que le tournedos ne soit cuit, saisir le foie gras. Poser le foie gras sur le tournedos,
garnir avec de fines tranches de truffes et servir immédiatement. Si vous n'avez pas de truffe, vous pouvez utiliser des morilles.
					\end{minipage}
					}
		}
		\\
		\subfloat[\wf associé.]{ \label{fig:exwf}
         		$
         			\miniwfconj
         				{\text{`frire le tournedos'}}
         				{\text{`saisir le foie gras'}}
         			\to
         			\begin{minipage}{50mm}
         				`poser le foie gras sur le tournedos'
         			\end{minipage}
         			\to
         			\miniwfdisj
         				{\text{`garnir de truffe'}}
         				{\text{`garnir de morille'}}
         			\to
         			\text{`servir'}
         		$
      }
	\end{center}
	\caption{Un exemple tiré du domaine culinaire}
	\label{fig:ex}
\end{figure*}

Les \wfs sont parfois spécifiés grâce à la définition de conditions portant sur l'exécution des activités, ce qui accroît leur expressivité mais aussi leur complexité. En particulier, les conditions sont exprimées dans un autre langage formel. La section~\ref{sec:concl} dessine quelques pistes à suivre pour gérer cet aspect, en utilisant des logiques classique et temporelle.

Par la suite, nous aurons besoin d'une fonction $S$ faisant correspondre à  un \wf  $\varphi$ un ensemble de \wfs  inclus dans $\varphi$, y compris $\varphi$.
Cette fonction peut être définie récursivement :
\begin{itemize}
        \item $\textit{S}(\varphi) = \textit{SP}(\varphi) \cup \{\varphi\}$;\hfill
	\item $\textit{SP}(\alpha) = \emptyset$;
	\item $\textit{SP}\left(\miniwfconj{\varphi}{\chi}\right)
             = \textit{SP}(\wfsequence{\varphi}{\chi})
             = \textit{SP}\left(\miniwfdisj{\varphi}{\chi}\right)
             = \textit{S}(\varphi) \cup \textit{S}(\chi)$;
	\item $\textit{SP}(\wfloop{\varphi}) = \textit{S}(\varphi)$\footnote{%
               En principe, il est aussi nécessaire de distinguer dans $\textit{S}(\varphi)$
                 les occurrences multiples du même \wf.
                Pour éluder cette difficulté  technique, nous supposons que,
                 \emph{modulo} renommage, chaque \wf  atomique n'apparaît qu'une fois dans un \wf,
                 e.g. $\textit{S}(\wfsequence{\wfpsequence{\alpha}{\beta} }{\alpha})
                     = \textit{S}(\wfsequence{\wfpsequence{\alpha_1}{\beta} }{\alpha_2})$.},
\end{itemize}
où  $\textit{SP}(\varphi)$ est l'ensemble des flux qui composent $\varphi$, hors $\varphi$ lui-même.

\section{Algèbres qualitatives temporelles}
\label{sec:qa}
Une algèbre qualitative est une algèbre de relations au sens de Tarski~\cite{tarski41jsl}. Elle est définie comme un ensemble fini  $\mathfrak{B}$ de relations binaires entre couples d'individus. Dans les algèbres qualitatives temporelles, les individus sont des points, des intervalles ou des sous-ensembles quelconques de la droite temporelle. Les relations sont définies de telle sorte que $\mathfrak{B}$ représente une partition de l'ensemble des couples d'individus (elles sont disjointes et exhaustives).

\begin{figure}
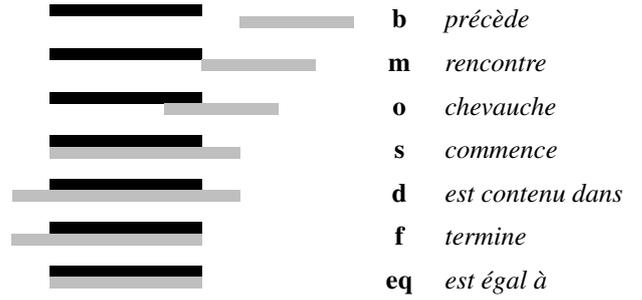

  {
  \def\intx{$\rule[1.5mm]{20mm}{1.5mm}$}
  \def\inty#1#2{{\textcolor{gray!50}{\hspace{#1mm}$\rule[0mm]{#2mm}{1.5mm}$}}}
  \def\ligne#1#2#3#4{\makebox[0mm][l]{\intx}\inty{#1}{#2} & $\textbf{#3}$ & \emph{#4}}
  \def\rien#1{#1} 
  %
  \begin{center}
    \begin{tabular}{l c l}
      \ligne{25}{15}{b}{précède}
      \\[1.5mm]
      \ligne{20}{15}{m}{rencontre}
      \\[1.5mm]
      \ligne{15}{15}{o}{chevauche}
      \\[1.5mm]
      \ligne{0}{25}{s}{commence}
      \\[1.5mm]
      \ligne{-5}{30}{d}{est contenu dans}
      \\[1.5mm]
      \ligne{-5}{25}{f}{termine}
      \\[1.5mm]
      \ligne{0}{20}{eq}{est égal à}
    \end{tabular}
  \end{center}
}

  \caption{Les relations de l'algèbre d'intervalles de Allen.
           \label{fig:allen}}
\end{figure}

L'algèbre qualitative la plus célèbre est l'algèbre d'intervalles de Allen~\cite{allen81ijcai,allen83cacm} qui définit un ensemble de 13 relations de base entre  couples d'intervalles, où chacune des relations correspond à une configuration possible des quatre extrémités de deux intervalles.  La figure~\ref{fig:allen} montre sept de ces relations, les six autres étant les relations inverses des six premières (la septième, \alrelnb{eq}, est symétrique).

Une algèbre plus simple est l'algèbre des points~\cite{vk86aaai}, où les individus sont des instants et l'ensemble des relations de base est  $\{<, =, >\}$. D'autres algèbres ont été proposées pour étendre l'algèbre de Allen. Un exemple intéressant est celui de   \indu\cite{pks99atai} qui étend l'ensemble des relations de Allen en les combinant avec trois relations,  $<$, $=$ et $>$, sur la durée des intervalles. Par exemple, $i \alrel{m$^<$} j$ signifie que la fin de $i$ coïncide avec le début de $j$ et que la durée de $i$ est plus petite que celle de $j$. Pour sept des relations d'Allen, il n'y a qu'une seule relation de durée possible (e.g. $i \alrel{d} j$ implique que la durée de $i$ est plus petite que la durée de $j$).  Pour les six autres, les trois relations de durée sont possibles. On obtient ainsi 25 relations de base.

Dans les algèbres qualitatives, les inférences peuvent être mises en {\oe}uvre en considérant les individus comme les n{\oe}uds d'un réseau de contraintes, dans le cadre de la résolution de problèmes de contraintes~\cite{mackworth77ai,montanari74is}. Dans les réseaux de contraintes qualitatives (RCQ) ainsi définis,   les arcs sont étiquetés par des sous-ensembles de l'ensemble  $\mathfrak{B}$ des relations de base, indiquant une disjonction de relations. Une des tâches réalisables dans un RCQ est la construction d'un scénario, i.e. la construction d'un sous-ensemble cohérent de relations entre les individus représentés dans le réseau. 

De manière formelle, un réseau de contraintes qualitatives  $\mathcal{N}$ est un couple  $(V,C)$, où $V$ est un ensemble de variables représentant des individus (objets temporels) et $C$ est un ensemble de contraintes de la forme $(V_i, V_j, C_{ij})$ avec $V_i, V_j \in V$ et $C_{ij} \in 2^\mathfrak{B}$. Un réseau est cohérent s'il est possible de trouver une valuation pour les variables de $V$ qui est telle qu'une relation de chaque  $C_{ij}$ soit satisfaite.

\begin{figure}
	\begin{center}
		\includegraphics[width=\columnwidth]{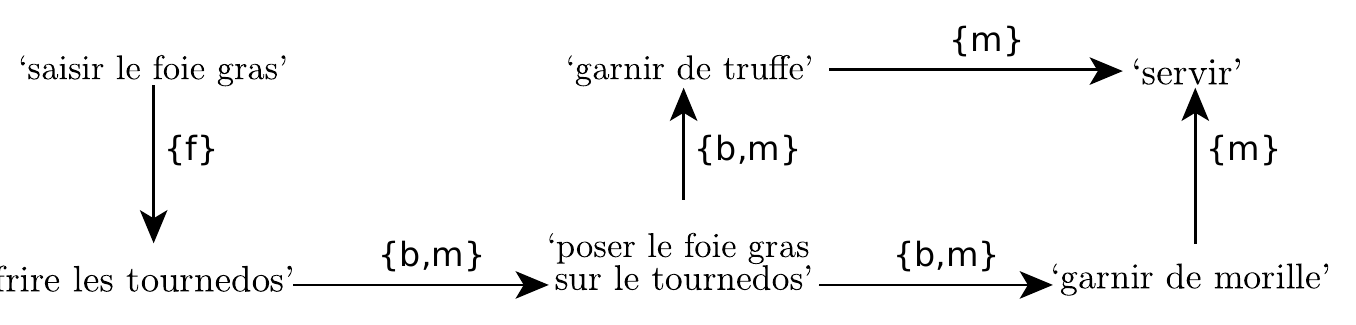}
		\caption{Un réseau de contraintes qualitatives représentant la recette~\ref{fig:extxt}.}
		\label{fig:exqa}
	\end{center}
\end{figure}

À notre avis, l'algèbre de relations \indu est la plus appropriée pour
représenter les relations temporelles entre et au sein d'activités,
mais comme l'algèbre des intervalles est plus simple, nous allons
l'utiliser pour illustrer notre démarche. Nous gardons toutefois à
l'esprit qu'\indu peut la remplacer facilement.
La figure~\ref{fig:exqa} montre une représentation de la recette de la  figure~\ref{fig:extxt} sous la forme d'un réseau de contraintes qualitatives utilisant l'algèbre d'intervalles de Allen : ce langage n'est pas suffisamment expressif pour rendre compte d'une disjonction avec plusieurs options. Représenter les relations entre ces options et les autres intervalles est également problématique, car une (au moins) des options ne sera pas réalisée.

\section{Une sémantique en théorie des modèles des flux opérationnels et des algèbres temporelles}
\label{sec:sem}

\subsection{Une sémantique des flux opérationnels}
\label{sec:wfsem}

Un flux opérationnel représente une définition de processus et
 chacune de ses exécutions donne une instance de processus.
Cette notion d'instance de processus va être utilisée pour
 définir une sémantique du formalisme des flux opérationnels.
L'idée est qu'un modèle d'un flux opérationnel est une exécution
 de ce flux.

L'interprétation d'un flux opérationnel dans un modèle donné est
 un événement, correspondant à une exécution de ce flux par un agent.
Le domaine d'interprétation est l'ensemble $\Omega$ des
 \emph{événements élémentaires}.
Un \emph{événement} est un sous-ensemble de $\Omega$.
L'ensemble vide est l'élément impossible et est la seule
 interprétation possible d'un flux opérationnel insatisfiable.

Un événement se déroule dans le temps.
Dans cet article, le temps est représenté de façon linéaire 
 par l'ensemble des rationnels, $\mathbb{Q}$,
 et chaque événement peut être mis en correspondance avec
 un sous-ensemble de $\mathbb{Q}$.

\paragraph{Interprétations.}
%
Une interprétation $\wfint$ est un triplet
 $(\Omega, \wfinterpretationfunction, \Theta)$ où:
\begin{itemize}
\item
  $\Omega$ est un ensemble non vide
   (d'événements élémentaires) ;
\item
  $\wfinterpretationfunction$ est une fonction associant
  à un flux opérationnel $\varphi \in \mathcal{W}$ un événement
  $\wfinterpretationof{\varphi} \in 2^\wfinterpretationdomain$ ;
\item
  $\wftimeofeventfunction$ est une fonction associant à un
  événement $E$ un sous-ensemble fermé et borné
  $\wftimeofevent{E}$ de $\mathbb{Q}$ tel que :
  \begin{itemize}
  \item
    $\wftimeofevent{\emptyset}=\emptyset$
    (i.e. l'événement impossible ne peut pas arriver) ;
  \item
    Si $E\neq\emptyset$,
     alors $\wftimeofevent{E}\neq\emptyset$
     (i.e. tout événement possible arrive effectivement) ;
  \item
    $\wftimeofevent{E \cup F} = \wftimeofevent{E} \cup \wftimeofevent{F}$
    pour $E, F \in 2^\wfinterpretationdomain$.
  \end{itemize}
\end{itemize}

Une interprétation est un modèle d'un flux opérationnel $\varphi$
 si elle associe à $\varphi$ un événement possible :
 $\wfinterpretationof{\varphi} \neq \emptyset$.

\paragraph{Conjonction.}
%
La conjonction de deux flux opérationnels représente une
 activité complexe sans contrainte particulière entre les
 termes de la conjonction.
Par conséquent, l'événement correspondant à la conjonction
 est l'union des événements de chaque terme, à moins qu'un
 des termes soit non satisfait par l'interprétation $\wfint$, auquel
 cas la conjonction n'est pas non plus satisfaite par $\wfint$.

\[
	\wfinterpretationof{\miniwfconj{\varphi}{\psi}} = 
	\begin{cases}
   	{ \wfinterpretationof{\varphi}\cup\wfinterpretationof{\psi} }&{
   		\tif \wfinterpretationof{\varphi} \neq \emptyset \tand \wfinterpretationof{\psi} \neq \emptyset
   	}\\{ \emptyset }&{ \totherwise }
	\end{cases}
\]

\paragraph{Sequence.}
%
La séquence de deux flux opérationnels est leur conjonction
 avec une contrainte additionnelle spécifiant que la première
 activité doit être finie avant que la deuxième commence.
\[
	\wfinterpretationof{(\wfsequence{\varphi}{\psi})} =
	\begin{cases}
     	 { \wfinterpretationof{\varphi}\cup\wfinterpretationof{\psi} }
   	&{ \tif \wfinterpretationof{\varphi} \neq \emptyset, \wfinterpretationof{\psi} \neq \emptyset }
   	\\
   	&{\tand \max\wftimeofevent{\wfinterpretationof{\varphi}} \leq \min\wftimeofevent{\wfinterpretationof{\psi}} }
   	\\
   	 { \emptyset }
   	&{ \totherwise }
	\end{cases}
\]

\paragraph{Disjonction.}
%
L'exécution d'une disjonction de deux flux opérationnels
 est l'exécution de l'un d'entre eux.

\[
	\wfinterpretationof{\miniwfdisj{\varphi}{\psi}} = \wfinterpretationof{\varphi}
	\tor
	\wfinterpretationof{\miniwfdisj{\varphi}{\psi}} = \wfinterpretationof{\psi}
\]

\paragraph{Boucle.}
%
Une boucle correspond à la répétition d'une activité
 (i.e. à une instanciation multiple et en séquence
  de plusieurs instances de la même définition de processus).
Soit $\wfrepetition{\varphi}{n} = ( \dots ( \wfsequence{\varphi}{\varphi} ) \wfsequenceoperator \dots \wfsequenceoperator \varphi )$, i.e.:
\begin{itemize}
\item
  $\wfrepetition{\varphi}{1} = \varphi$
\item
  $\wfrepetition{\varphi}{(n+1)} = \wfsequence{ ( \wfrepetition{\varphi}{n} ) }{ \varphi }$
\end{itemize}
On peut, sur la base de cette notation, définir la sémantique de la boucle comme suit :
\[
	\text{Il existe $n \geq 1$ tel que} \qquad
	\wfinterpretationof{ \wfloop{\varphi} } = \wfinterpretationof{\wfrepetition{\varphi}{n}}
\]

\paragraph{Inférences.} 
%
Trois inférences déductives sont considérées :
 la vérification de la satisfiabilité, de la subsomption et de l'équivalence.
Un flux opérationnel est dit satisfiable s'il possède un modèle.
Un flux opérationnel $\varphi$ est subsumé par un flux opérationnel
 $\chi$ ($\varphi\subsumepar\chi$) si
 toute exécution de $\varphi$ est une exécution de $\chi$ :
 pour toute interprétation $\wfint$, si $\wfint$ est un modèle
 de $\varphi$ alors $\wfint$ est un modèle de $\chi$,
 et $\wfinterpretationof{\varphi}=\wfinterpretationof{\chi}$.
Deux flux opérationnels $\varphi$ et $\chi$ sont équivalents
 ($\varphi\equivalenta\chi$)
 si $\varphi\subsumepar\chi$ et $\chi\subsumepar\varphi$.
$\subsumepar$ est un préordre sur $\langagewf$
 et $\equivalenta$ est une relation d'équivalence sur $\langagewf$.

\paragraph{Propriétés.} 
%
Tout flux opérationnel $\varphi\in\mathcal{W}$ est satisfiable\footnote{%
  On peut montrer cela par récurrence sur la taille d'un
   flux opérationnel où cette taille est définie par
   $t=a+c$ avec $a$ le nombre de concepts atomiques
   apparaissant dans le flux opérationnel et $c$
   le nombre de constructeurs qu'il contient
   --- p. ex., $t=3$ pour $\wfsequence{\alpha}{\beta}$.}.
Il faut noter que quand un flux opérationnel atomique apparaît
 plusieurs fois dans un flux opérationnel, chacune de ses
 occurrences est interprétée indépendamment des autres
 (ce qui permet, par exemple, que $\wfsequence{\alpha}{\alpha}$
  soit satisfiable).
Cette propriété est vérifiée pour les flux opérationnels dans
 le formalisme $\langagewf$ qui empêche,
 par exemple, la représentation de flux opérationnels
 insatisfiables tels que
 \raisebox{-3mm}{{
  \def\leve#1#2{\raisebox{#1mm}{#2}}
  \def\mef#1{{\huge$#1$}} 
  \def\aa{\mef{\alpha}}
  \def\bb{\mef{\beta}}
  \!\!\!\!\!\!\!\leve{0}{\resizebox{0.08\textwidth}{!}{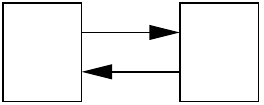}}\!\!\!
}
}.

Modulo l'équivalence,
 la disjonction et la boucle sont idempotententes,
 la conjonction et la disjonction sont commutatives et
 associatives et la disjonction est autodistributive.
Autrement écrit, pour $\varphi, \chi, \psi\in\langagewf$ :
%
 \begin{gather*}
    \wfdisj{\varphi}{\varphi}\equivalenta\varphi
    \qquad
    \wfloop{\wfloop{\varphi}}\equivalenta\wfloop{\varphi}
    \qquad
    \wfconj{\varphi}{\chi}\equivalenta\wfconj{\chi}{\varphi}
    \qquad
    \wfdisj{\varphi}{\chi}\equivalenta\wfdisj{\chi}{\varphi}
    \\
    \wfconj{\wfconj{\varphi}{\chi}}{\psi}\equivalenta\wfconj{\varphi}{\wfconj{\chi}{\psi}}
    \qquad
    \wfdisj{\wfdisj{\varphi}{\chi}}{\psi}\equivalenta\wfdisj{\varphi}{\wfdisj{\chi}{\psi}}
    \qquad
    \wfdisj{\varphi}{\wfdisj{\chi}{\psi}} \equivalenta
    \wfdisjIIsepar{\wfdisj{\varphi}{\chi}}{\wfdisj{\varphi}{\psi}}
 \end{gather*}
 (les preuves sont immédiates).
Ces équivalences rendent possible la simplification de conjonctions
 (resp., disjonctions) emboîtées :
 elles sont équivalentes à des conjonctions (resp., disjonctions)
 <<\,plates\,>> de multiensembles (resp., ensembles)
 de flux opérationnels.
Par exemple :
 \begin{equation*}
   \wfconjIIsepar{\wfconj{\alpha}{\beta}}{\wfconj{\alpha}{\gamma}}
   \equivalenta
   \wfconjIV{\alpha}{\alpha}{\beta}{\gamma}
   \qquad\qquad
   \wfdisjIIsepar{\wfdisj{\alpha}{\beta}}{\wfdisj{\alpha}{\gamma}}
   \equivalenta
   \wfdisjIII{\alpha}{\beta}{\gamma}
 \end{equation*}

Le fait qu'une boucle soit une séquence itérée implique
 les équivalences et subsomptions suivantes :
 \begin{gather*}
   \wfsequence{\varphi}{\wfloop{\varphi}} \equivalenta \wfsequence{\wfloop{\varphi}}{\varphi}
   \qquad
   \wfsequence{\wfloop{\varphi}}{\wfloop{\varphi}} \equivalenta \wfsequence{\wfloop{\varphi}}{\varphi}
   \\
   \varphi \subsumepar \wfloop{\varphi}
   \qquad
   \wfsequence{\wfloop{\varphi}}{\varphi} \subsumepar \wfloop{\varphi}
 \end{gather*}

Notons également les propriétés suivantes :
 \begin{gather*}
   \text{pour $\varphi, \chi\in\langagewf$,}
   \qquad
   \varphi \subsumepar \wfdisj{\varphi}{\chi}
   \qquad
   \wfsequence{\varphi}{\chi} \subsumepar \wfconj{\varphi}{\chi}
 \end{gather*}

Finalement, si $\varphi$, un sous-flux de $\Phi$
 (au sens de $\varphi\in{}S(\Phi)$),
 est généralisé (au sens de $\subsumepar$),
 alors $\Phi$ est généralisé.
Formellement, cela signifie que si $\Phi, \Psi, \varphi, \psi\in\langagewf$
 est tel que $\varphi\in{}S(\Phi)$, $\varphi\subsumepar\psi$
 et $\Psi$ est obtenu en substituant (une occurrence de) $\varphi$
 par $\psi$ dans $\Phi$, alors $\Phi\subsumepar\Psi$.
Cela peut être prouvé en prouvant d'abord les lemmes suivants,
 pour $\varphi, \chi, \psi\in\langagewf$ tels que $\varphi\subsumepar\psi$ :
 \begin{equation*}
   \wfconj{\varphi}{\chi} \subsumepar \wfconj{\psi}{\chi}
   \qquad
   \wfsequence{\varphi}{\chi} \subsumepar \wfsequence{\psi}{\chi}
   \qquad
   \wfdisj{\varphi}{\chi} \subsumepar \wfdisj{\psi}{\chi}
   \qquad
   \wfloop{\varphi} \subsumepar \wfloop{\psi}
 \end{equation*}
Puis le résultat est prouvé par récurrence sur la profondeur de
 $\varphi$ dans $\Phi$.

\subsection{Une sémantique pour les algèbres d'intervalles}
%
\label{sec:qasem}

Les algèbres d'intervalles expriment des relations temporelles dans des
 couples d'intervalles formels.
Les intervalles \emph{formels} (i.e. éléments du formalisme)
 sont généralement interprétés comme étant des
 intervalles \emph{numériques} de $\mathbb{Q}$,
 et les relations entre deux intervalles numériques peuvent être
 définies à l'aide de relations entre leurs bornes, comme défini
 dans~\cite{allen81ijcai}.
En interprétant les intervalles formels par des intervalles numériques,
 il est facile de réutiliser ces définitions pour proposer une
 sémantique.

\paragraph{Interprétations.}
%
Une interprétation $\qaint$ est composée de la fonction 
 $\induinterpretationfunction$ associant à un intervalle formel $i$
 un intervalle numérique fermé, borné et non réduit à un point.
Une interprétation $\qaint$ est un modèle d'un réseau de contraintes
 qualitatives $\mathcal{N}=(V,C)$ s'il satisfait toutes ses contraintes :
\[
	\qaint \models \mathcal{N} \, \text{ si, pour tout } (V_i,V_j,C_{ij}) \in C, \, \qaint \models V_i \mathrel{C_{ij}} V_j
\]

La sémantique des relations est définie comme suit :
\begin{raggedright}
\begin{itemize}
\item $\displaystyle \qaint \models i \mathrel{R} j \tif \qaint \models i \mathrel{\{r\}} j \text{ pour un } r \in R$ ;
\item $\qaint \models i \alrel{b} j \text{ si } \max \induinterpretationof{i} < \min \induinterpretationof{j}$ ;
\item $\qaint \models i \alrel{m} j \text{ si } \max \induinterpretationof{i} = \min \induinterpretationof{j}$ ;
\item 
  $
  \qaint \models i \alrel{o} j \text{ si } \min \induinterpretationof{i} < \min \induinterpretationof{j},$
  $\max \induinterpretationof{i} > \min \induinterpretationof{j}$
  et
  $\max \induinterpretationof{i} < \max \induinterpretationof{j}$ ;
\item
  $\qaint \models i \alrel{s} j$ si
  $\min \induinterpretationof{i} = \min \induinterpretationof{j}$
  et \linebreak[4] $\max \induinterpretationof{i} < \max \induinterpretationof{j}$ ;
\item
  $\qaint \models i \alrel{d} j$
  si $\min \induinterpretationof{i} > \min \induinterpretationof{j}$
  et \linebreak[4] $\max \induinterpretationof{i} < \max \induinterpretationof{j}$ ;
\item
  $\qaint \models i \alrel{f} j$
  si $\min \induinterpretationof{i} > \min \induinterpretationof{j}$
  et \linebreak[4] $\max \induinterpretationof{i} = \max \induinterpretationof{j}$ ;
\item
  $\qaint \models i \alrel{eq} j$
  si $\min \induinterpretationof{i} = \min \induinterpretationof{j}$
  et \linebreak[4] $\max \induinterpretationof{i} = \max \induinterpretationof{j}$ ;
\item
  $\qaint \models i \mathrel{ \{ r\qainv \}} j$
  si $\qaint \models j \mathrel{\{r\}} i$,
  avec $r\qainv$ défini par
  $\alrelnb{ p\qainv }=\alrelnb{ pi}$,
  $\alrelnb{ m\qainv }=\alrelnb{ mi}$,
  $\alrelnb{ o\qainv }=\alrelnb{ oi}$,
  $\alrelnb{ s\qainv }=\alrelnb{ si}$,
  $\alrelnb{ d\qainv }=\alrelnb{ di}$ et
  $\alrelnb{f\qainv } = \alrelnb{ fi}$.
\end{itemize}
\end{raggedright}

\paragraph{Inférences.} 
%
Trois inférences de base sont considérées :
 la vérification de la satisfiabilité (ou cohérence), de l'implication
 et de l'équivalence.
Un RCQ $\QCN$ est cohérent s'il a au moins une solution,
 c'est-à-dire une instanciation de ses variables telle que
 toutes ses contraintes sont satisfaites.
Une telle solution est équivalente à un modèle de $\mathcal{N}$.
Un RCQ $\QCN_1$ entraîne un RCQ $\QCN_2$ ($\QCN_1\models\QCN_2$)
 si tout modèle du premier est un modèle du second.
$\QCN_1$ est équivalent à $\QCN_2$ ($\QCN_1\equiv\QCN_2$)
 si $\QCN_1\models\QCN_2$ et $\QCN_2\models\QCN_1$.

%

Tester la cohérence d'un RCQ se fait généralement à l'aide d'un
 algorithme de propagation de contraintes~\cite{allen83cacm}
 s'appuyant sur la cohérence de chemin et sur une table de
 composition des relations~\cite{sv98c}.
Ce test de cohérence est NP-complet~\cite{vkb89rqrps}.
La table de composition indique, par exemple, que la
  composition de $\alrel{f}$ et $\alrel{m}$ donne
 $\alrel{m}$, ce qui signifie que si $i\alrel{f}j$ et $j\alrel{m}k$
 alors $i\alrel{m}k$.

\section{Extension des flux opérationnels par une algèbre temporelle}
\label{sec:ext}
Ainsi que cela a été discuté en introduction, ni les
 flux opérationnels tels que définis actuellement, ni
 les algèbres qualitatives, n'ont un pouvoir expressif
 suffisant pour représenter pleinement toutes les relations
 temporelles possibles dans l'exécution de tâches complexes.
L'expressivité temporelle des flux opérationnels est limitée
 à une contrainte de précédence (exprimée dans les séquences)
 alors que les algèbres qualitatives sont limitées par leur
 absence d'alternatives et de répétitions (représentées
 par les disjonctions et les boucles des flux opérationnels).

Cet article suggère d'annoter les flux opérationnels par des
 intervalles formels de l'algèbre de Allen.
Dans la section~\ref{sec:wfsem}, nous avions défini la fonction
 $\Theta$ associant aux événements des flux opérationnels des
 sous-ensembles de $\mathbb{Q}$.
La fermeture convexe de chacun des ces sous-ensembles est un
 intervalle numérique, comme l'est chaque interprétation
 d'un intervalle formel, comme définie dans la
 section~\ref{sec:qasem}.
Par conséquent, il est possible d'associer aux flux opérationnels
 des intervalles formels, de façon à spécifier des contraintes
 qualitatives sur leurs temps d'exécution.

\subsection{Syntaxe}

\begin{figure*}
	\begin{center}
		$
			         \miniwfconj
         				{\text{frire le tournedos}}
         				{\text{saisir le foie gras}}
         			\to
         			\begin{minipage}{55mm} 
         				poser le foie gras sur le tournedos 
         			\end{minipage}
         			\to
         			\miniwfdisj
         				{\text{garnir de truffe}}
         				{\text{garnir de morille}}
         			\to
         			\text{servir}
                $\\
         	\includegraphics[width=.75\textwidth]{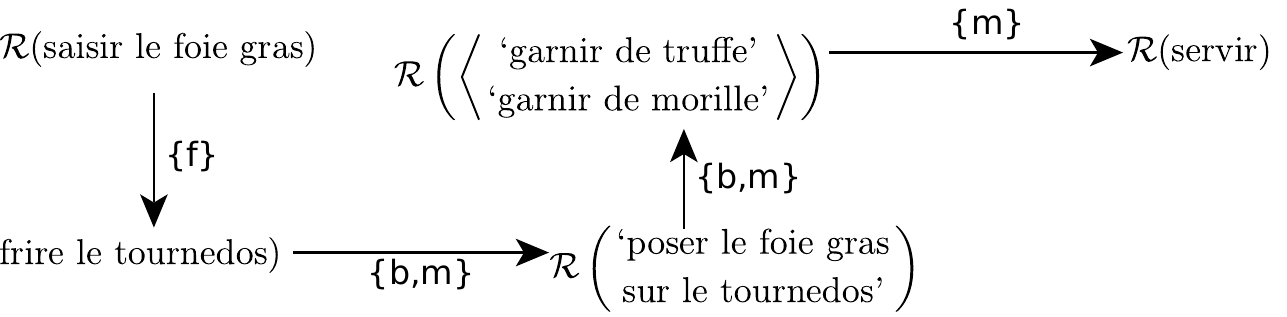}
	\caption{Flux opérationnel étendu associé à la recette~\ref{fig:extxt}.\label{fig:exewf}}
	\end{center}
\end{figure*}

Le langage des flux opérationnels étendus, $\mathcal{W}_\mathfrak{B}$,
 est une combinaison du langage des flux opérationnels, $\mathcal{W}$,
 et d'une algèbre qualitative  $\mathfrak{B}$ sur les intervalles,
 p. ex., l'algèbre de Allen.
Si $\varphi$ est un flux opérationnel
 (avec $S(\varphi)$ l'ensemble des flux opérationnels
  qu'il contient) et $\QCN=(V,C)$ est un réseau de contraintes
 qualitatives, alors $(\varphi,\QCN)$ est un flux opérationnel étendu.
On se donne une injection $\wfqcninject$ de $S(\varphi)$ dans $V$
 telle que l'intervalle correspondant à $\varphi$ s'écrit
 $\wfqcninject(\varphi)$.
La figure~\ref{fig:exewf} présente le flux opérationnel représentant
 la recette de la figure~\ref{fig:extxt}, combinant
 le pouvoir expressif des flux opérationnels et des algèbres
 qualitatives.

Dans ce formalisme, on interdit d'exprimer des relations entre
 deux intervalles formels correspondant l'un à un flux opérationnel
 à l'intérieur d'une boucle et l'autre à un flux opérationnel à
 l'extérieur de cette boucle.
Cela ne limite pas l'expressivité car il est possible d'exprimer
 la relation à l'extérieur de la boucle.
Ainsi, dans $\wfsequence{\alpha}{\wfloop{\beta}}$,
 une relation peut être exprimée entre
 $\wfqcninject(\alpha)$ et $\wfqcninject(\wfloop{\beta})$
 mais pas entre $\wfqcninject(\alpha)$ et $\wfqcninject(\beta)$.

\subsection{Sémantique}

Les sémantiques décrites aux sections~\ref{sec:wfsem} et~\ref{sec:qasem}
 peuvent être combinées.

Une interprétation $\ewfint$ d'un flux opérationnel étendu est un
 triplet $(\Omega, \extwfinterpretationfunction, \Theta)$, où :
\begin{itemize}
\item
  $\Omega$ est un ensemble non vide ;
 \item
    $\wftimeofeventfunction$ est une fonction associant à
     un événement $E \subseteq \Omega$ un ensemble fermé
     et borné $\wftimeofevent{E} \subseteq \mathbb{Q}$ tel que :
     \begin{itemize}
     \item
       $\wftimeofevent{\emptyset}=\emptyset$;
     \item
       Si $E\neq\emptyset$ alors $\wftimeofevent{E}\neq\emptyset$ ;
     \item
       $\wftimeofevent{E \cup F} = \wftimeofevent{E} \cup \wftimeofevent{F}$
        pour tout $E, F \in 2^\wfinterpretationdomain$ ;
     \end{itemize}
\item
  $\wfinterpretationfunction$ est une fonction qui
  \begin{itemize}
  \item
    à un flux opérationnel (classique) $\varphi\in\langagewf$
    associe un événement $\wfinterpretationof{\varphi} \in 2^\wfinterpretationdomain$ ;
  \item
    à un intervalle formel $i$
    associe $\induinterpretationof{i} \subseteq \intervaldomain$,
    un intervalle numérique fermé, borné et non réduit à un point ;
  \item
    à un flux opérationnel étendu $(\varphi, \QCN)\in\langagewfe$
    associe un événement $\wfinterpretationof{(\varphi, \QCN)}\in2^\wfinterpretationdomain$
    tel que :
   \begin{equation*}
     \wfinterpretationof{(\varphi, \QCN)}
     =
     \begin{cases}
     \wfinterpretationof{\varphi}
     & \text{si }
       \wfinterpretationof{\varphi}\neq\emptyset,
       \interpretation\models\QCN
       \text{ et}
     \\
     & \text{\phantom{si }pour tout }
       \psi\in{}S(\varphi),
     \\
     & \text{\phantom{si pour tout }}
     \widehat{\Theta(\wfinterpretationof{\psi})} = \induinterpretationof{\wfqcninject(\psi)}
     \\
     \emptyset
     & \text{sinon}
     \end{cases}
   \end{equation*}
   où $\widehat{\vartheta}$ est la fermeture convexe du sous-ensemble $\vartheta$ de $\mathbb{Q}$
   (i.e. le plus petit intervalle numérique contenant $\vartheta$).
   \end{itemize}
\end{itemize}



Un flux opérationnel $\varphi\in\langagewf$ est assimilé
 à un flux opérationnel étendu $(\varphi, \emptyset)$
 où $\emptyset$ dénote le RCQ vide.
$\langagewf$ est ainsi plongé dans $\langagewfe$.

\subsection{Propriétés}

$(\varphi, \QCN)$ est satisfiable s'il a un modèle.
$(\varphi_1, \QCN_1)\subsumepar(\varphi_2, \QCN_2)$ si, pour
 tout modèle $\interpretation$ de $(\varphi_1, \QCN_1)$,
 $\interpretation$ est un modèle de $(\varphi_2, \QCN_2)$
 et $\wfinterpretationof{\varphi_1}=\wfinterpretationof{\varphi_2}$.
$(\varphi_1, \QCN_1)\equivalenta(\varphi_2, \QCN_2)$ si
 $(\varphi_1, \QCN_1)\subsumepar(\varphi_2, \QCN_2)$ et
 $(\varphi_2, \QCN_2)\subsumepar(\varphi_1, \QCN_1)$.
Ces inférences étendent les inférences de $\langagewf$ :
 $(\varphi, \emptyset)$ est satisfiable ssi $\varphi$ est satisfiable (i.e. toujours) ;
 $(\varphi_1, \emptyset)\subsumepar(\varphi_2, \emptyset)$ ssi $\varphi_1\subsumepar\varphi_2$;
 $(\varphi_1, \emptyset)\equivalenta(\varphi_2, \emptyset)$ ssi $\varphi_1\equivalenta\varphi_2$.

Pour tout $(\varphi, \QCN) \in \langagewfe$, il existe
 $(\varphi', \QCN') \in\langagewfe$ qui est équivalent
 et tel que $\varphi'$ ne contient pas de séquence.
$(\varphi', \QCN')$ est dit sans séquence.
Cela peut être montré grâce à l'équivalence
 $(\varphi, \QCN)\equivalenta(\varphi_1, \QCN_1)$ où
 $(\varphi, \QCN)$ est un flux opérationnel étendu
 tel qu'il existe $\wfsequence{\chi}{\psi}\in{}S(\varphi)$,
 $\varphi_1$ est obtenu en remplaçant $\wfsequence{\chi}{\psi}$ par
 $\wfconj{\chi}{\psi}$ et
 $\QCN_1=\QCN\cup\{\lien(\chi)\alrel{b, m}\lien(\psi)\}$ :
 l'application répétée de gauche à droite de cette équivalence
 finit par donner un flux opérationnel étendu sans séquence
 appelé \emph{le} flux opérationnel sans séquence associé à
 $(\varphi, \QCN)$.
Par exemple :
{
\def\ns{\!\!\!\!}
 \begin{equation*}
   (\wfsequenceIII{\alpha}{\beta}{\gamma}, \emptyset)
   \equivalenta
   \left(\wfconjIII{\alpha}{\beta}{\gamma},
   \left\{\mlm{\ns$\lien(\alpha)\alrel{b, m}\lien(\beta),$\ns \\
               \ns$\lien(\beta)\alrel{b, m}\lien(\gamma)$\ns}\right\}\right)
 \end{equation*}
}

Soit $(\varphi, \QCN)$ un flux opérationnel étendu sans séquence.
Si $\QCN$ est satisfiable alors $(\varphi, \QCN)$ est satisfiable.
La réciproque n'est pas vraie en général, comme le contre-exemple
 suivant le montre :
 \begin{gather*}
   (\varphi, \QCN)
   = \left(\wfconjIII{\wfdisj{\alpha}{\beta}}{\gamma}{\delta},
           \raisebox{-8mm}{\begin{tikzpicture}[node distance=1.5cm]
  \node(alpha)[]{$\lien(\alpha)$} ;
  \node(sousalpha)[below of=alpha]{} ;
  \node(gamma)[right of=sousalpha]{$\lien(\gamma)$} ;
  \node(delta)[left of=sousalpha]{$\lien(\delta)$} ;
  \draw[->] (alpha) -- (gamma) node[midway, above]{$~\alrel{b}$};
  \draw[->] (gamma) -- (delta) node[midway, above]{$\alrel{b}$};
  \draw[->] (delta) -- (alpha) node[midway, above]{$\alrel{b}~$};
\end{tikzpicture}}
   \!\!\!\!\!\right)
   \\
   \text{$(\varphi, \QCN)$ est satisfiable bien que $\QCN$ ne le soit pas.}
 \end{gather*}
Cependant, nous considérons que manipuler des flux opérationnels étendus
 $(\varphi, \QCN)$ tels que $\QCN$ est insatisfiable n'est pas souhaitable,
 d'où l'introduction de la notion de satisfiabilité forte :
 $(\varphi, \QCN)\in\langagewfe$ est fortement satisfiable si
 $\QCN'$ est satisfiable, où $(\varphi', \QCN')$ est le flux opérationnel
 sans séquence associé à $(\varphi, \QCN)$.
La satisfiabilité forte entraîne la satisfiabilité.
Puisque la mise sous forme sans séquence est polynomiale
 et que la satisfiabilité d'un réseau de contraintes qualitatives est
 NP-complète, 
 la satisfiabilité forte d'un flux opérationnel est également
 NP-complète.

Soit $(\varphi_1, \QCN_1)$ et $(\varphi_2, \QCN_2)$ deux flux opérationnels étendus.
Si $\varphi_1\subsumepar\varphi_2$ et $\QCN_1\models\QCN_2$ alors
 $(\varphi_1, \QCN_1)\subsumepar(\varphi_2, \QCN_2)$.




\section{Discussion et travaux connexes}
\label{sec:disc}
De nombreux travaux se sont déjà attachés à proposer une base formelle correcte pour les \wfs. Nous présentons rapidement certains de ces formalismes puis nous discutons en détail l'utilisation des réseaux de Pétri pour ce même objectif. Nous analysons ensuite les points faibles des algèbres qualitatives qui les rendent  insuffisantes pour représenter des processus et discutons de quelques approches qui pourraient compenser ces points faibles.

\paragraph{Formalismes comparables.}

Le premier objectif des travaux portant sur la définition d'une formalisation des \wfs est de fournir une sémantique exécutable pour aider au fonctionnement des systèmes de gestion de \wfs ainsi qu'un mécanisme de validation (vérification de cohérence). La plupart des travaux s'appuient sur les algèbres de processus car, comme ces algèbres sont précisément développées pour décrire des processus reliés, elles sont parfaitement adaptées pour représenter des \wfs.

Un exemple connu utilise le $\pi$-calcul~\cite{pw05bpm}, tandis que d'autres travaux plus récents ont étudié l'utilisation du langage des \emph{Communicating Sequential Processes}, développé pour décrire des systèmes concurrents~\cite{wong07sc}. Une alternative, l'utilisation de \emph{statecharts}, un formalisme très proche des \wfs défini pour les systèmes réactifs, est proposée dans~\cite{ww97icdt}. Bien que tous ces modèles proposent de solides fondations pour les \wfs, il semble difficile d'y intégrer des contraintes temporelles.

\paragraph{\Wfs et réseaux  de Petri.}

Les \wfs ont longtemps été plus ou moins associés aux réseaux de Pétri, mais la première modélisation sérieuse ne date que de 1993~\cite{en93atpn}. Le cadre le plus utilisé aujourd'hui est celui des réseaux de \wfs, décrits dans~\cite{aalst96tr}. 

Les réseaux de Pétri sont intéressants car ils possèdent de nombreuses extensions, en particulier pour représenter des connaissances temporelles plus fines, qui peuvent alors être intégrées dans les modèles de \wfs, par le biais de la notion de réseaux de \wfs temporisés (voir notamment~\cite{ls00smc} et~\cite{tm05snasc}).

Des durées, ou des intervalles de temps pendant lesquels le franchissement est possible, sont associés aux transitions. Cela permet de bien exprimer des relations temporelles entre les activités si la durée de chaque activité est connue précisément (et non variable d'une exécution à l'autre).  Dans les autres cas cependant, il semble qu'on perde certaines ou toutes les relations entre les moments où ont lieu les activités, relations qui seraient exprimables dans l'algèbre d'intervalles de Allen.

En revanche, le langage de \wfs étendus $\mathcal{W}_\indu$, qui utilise l'algèbre temporelle \indu (ajoutant ainsi la possibilité d'exprimer les durées des activités), rend possible d'exprimer à la fois les contraintes des réseaux de \wfs temporisés et les relations de Allen.

De plus, la définition d'algorithmes pour vérifier la cohérence des  réseaux de \wfs temporisés est un problème non résolu pour un certain nombre de formalismes, voir par exemple~\cite{tm09smc}.
En revanche la satisfiabilité des \wfs étendus peut être calculée grâce aux algorithmes classiques de satisfiabilité pour les RCQ.

\paragraph{Algèbres qualitatives.}

Afin de renforcer l'efficacité calculatoire de son modèle, Allen a explicitement interdit la disjonction, sauf pour la disjonction de relations entre deux intervalles~\cite{allen83cacm}.  \`A notre connaissance, aucun progrès n'a été réalisé sur cet aspect. De ce fait, la seule manière de raisonner sur un réseau contenant des disjonctions d'intervalles serait de le représenter sous une forme normale disjonctive --- soit, sous la forme d'une disjonction de réseaux ---  et d'appliquer les algorithmes existants sur chaque réseau normalisé, générant ainsi une croissance exponentielle en rapport avec le nombre de disjonctions dans le réseau d'origine.

Les boucles, quant à elles,  pourraient être représentées en utilisant des intervalles cycliques comme proposé par~\cite{osmani99mais} mais dans ce cas, des relations fondamentales comme la précédence perdent toute signification. Si nous voulons affecter à chaque activité distincte d'une boucle un intervalle correspondant à la fermeture convexe de ses occurrences, nous pouvons utiliser le modèle des intervalles généralisés proposé dans~\cite{ligozat91aaai} et ainsi représenter les relations entre les instances des différentes activités inclues dans la boucle.

\section{Conclusion et perspectives}
\label{sec:concl}
Cet article présente un formalisme de flux opérationnels étendus
 intégrant une algèbre qualitative.
À cette fin, une syntaxe a été définie pour les flux opérationnels
 et une sémantique commune aux flux opérationnels et aux algèbres
 qualitatives d'intervalles a été proposée.
Cela a rendu possible non seulement de proposer un formalisme
 comprenant à la fois des flux opérationnels et des algèbres
 d'intervalles, mais aussi de décrire des propriétés et des
 inférences sur les flux opérationnels et les flux opérationnels
 étendus.

Il faut garder à l'esprit que le but principal du papier était
 d'établir un lien entre les deux formalismes qui puisse être
 utilisé afin de représenter des connaissances de natures
 similaires dans deux types de formalismes,
 celui des flux opérationnels et celui des algèbres qualitatives.
Le but premier n'était donc pas tant de créer un nouveau formalisme
 que de créer un langage intermédiaire, fondé sur une sémantique
 commune, afin de permettre la communication entre un système
 utilisant des flux opérationnels et un système raisonnant sur
 des algèbres qualitatives temporelles.

Cela explique que les questions algorithmiques n'étaient pas
 centrales dans cet article.
Une première perspective est de se pencher sur ces questions,
 en particulier sur l'étude de la complexité des inférences et
 sur la conception d'algorithmes efficaces.
À terme, nous envisageons de développer un moteur d'inférences
 fondé sur le formalisme étendu.
Nous voulons aussi examiner comment deux systèmes à base de connaissances
 construits l'un sur des flux opérationnels classiques et l'autre sur
 une algèbre qualitative pourrait échanger des connaissances
 (notamment le résultat de leurs inférences) en s'appuyant sur
 cette sémantique commune.

Par ailleurs, il devrait être possible d'étendre la sémantique proposée,
 et ainsi, le cadre, à d'autres formalismes décrivant des connaissances
 similaires sur les processus.
ConGolog~\cite{degiacomo00ai} est une application intéressante,
 candidate pour cette étude.

Comme indiqué à la section~\ref{sec:wf},
 les disjonctions et les boucles peuvent dépendre de conditions.
Ce travail, comme beaucoup d'autres travaux dans ce domaine, a
 choisi d'ignorer cela provisoirement, mais il pourrait être
 étendu facilement, par exemple en utilisant une logique classique
 $\mathcal{L}$, comme la logique propositionnelle,
 ce qui conduirait au langage $\mathcal{W}_{\mathfrak{B},\mathcal{L}}$.
Un nouveau constructeur, la disjonction conditionnelle, serait
 introduit de la façon suivante : si $\varphi$ et $\psi$ sont
 deux flux opérationnels étendus et $p$ est une proposition
 de $\mathcal{L}$, $\miniwfcdisj{p}{\varphi}{\psi}$ est une
 disjonction conditionnelle.
Intuitivement, si $p$ est vrai, le flux opérationnel $\varphi$ est
 exécuté, sinon, c'est $\psi$.
La représentation de la recette en figure~\ref{fig:exewf} pourrait être
 complétée en remplaçant
 $\miniwfdisj{\text{garnir de truffe}}{\text{garnir de morille}}$ par
 $\miniwfcdisj{\text{a\_des\_truffes}}{\text{garnir de truffe}}{\text{garnir de morille}}$.
La sémantique pourrait être étendue comme suit :
\begin{itemize}
\item
  La fonction d'interprétation pourrait être étendue
   sur les propositions : $p\in\mathcal{L}$
   aurait pour image une valeur de $\proptruthvalues$ ;
\item
  $\wfinterpretationof{\miniwfcdisj{p}{\varphi}{\psi}} =
			\conddef{ \wfinterpretationof{\varphi} }{ \tif \wfinterpretationof{p} = \proptrue }{ \wfinterpretationof{\psi} }{ \totherwise }
		$
\end{itemize}

Une logique classique ne serait pas suffisante pour
 les conditions des boucles, puisque la même expression
 conditionnelle devra être évaluée avec des valeurs
 différentes, pour permettre d'arrêter la boucle.
Certaines logiques temporelles pourraient être adéquates
 pour exprimer des conditions de boucle, mais un mécanisme
 doit être défini pour permettre leur interaction avec les
 activités des flux opérationnels.

\bibliography{./biblio} 

\end{document}